%% file: icml2026/main.tex
\documentclass{article}

\usepackage{microtype}
\usepackage{graphicx}
\usepackage{subcaption}
\usepackage{booktabs} 
\usepackage{multirow}

\usepackage{hyperref}

\usepackage[preprint]{icml2026}

\usepackage{amsmath}
\usepackage{amssymb}
\usepackage{mathtools}
\usepackage{amsthm}

\usepackage[capitalize,noabbrev]{cleveref}

\usepackage{pifont}

\theoremstyle{plain}

\theoremstyle{definition}

\theoremstyle{remark}

\usepackage[textsize=tiny]{todonotes}

\icmltitlerunning{Robust Skills, Brittle Grounding: Diagnosing Restricted Generalization in VLA Policies }

\begin{document}

\twocolumn[
  \icmltitle{Robust Skills, Brittle Grounding: Diagnosing Restricted Generalization in Vision-Language Action Policies via Multi-Object Picking}



  \icmlsetsymbol{equal}{*}

  \begin{icmlauthorlist}
    \icmlauthor{David Emukpere}{equal,comp}
    \icmlauthor{Romain Deffayet}{equal,comp}
    \icmlauthor{Jean-Michel Renders}{comp}
  \end{icmlauthorlist}

  \icmlaffiliation{comp}{Naver Labs Europe, 6 Chem. de Maupertuis, Meylan, France}

  \icmlcorrespondingauthor{David Emukpere}{david.emukpere@naverlabs.com}
  \icmlcorrespondingauthor{Romain Deffayet}{romain.deffayet@naverlabs.com}

  \icmlkeywords{VLA, Generalization}

  \vskip 0.3in
]



\printAffiliationsAndNotice{\icmlEqualContribution}

\input{icml2026/0_abstract}
\input{icml2026/1_introduction}

\input{icml2026/2_related}

\input{icml2026/3_method}

\input{icml2026/4_experiments}

\input{icml2026/5_discussion}
\input{icml2026/6_conclusion}


\bibliography{icml2026/references}
\bibliographystyle{icml2026}

\newpage
\appendix
\onecolumn

\input{icml2026/7_appendix}


\end{document}

%% file: icml2026/0_abstract.tex
\begin{abstract}
Vision-language action (VLA) policies often report strong manipulation benchmark performance with relatively few demonstrations, but it remains unclear whether this reflects robust language-to-object grounding or reliance on object--location correlations that do not transfer beyond the training distribution. We present a controlled multi-object picking study that progressively increases object placement variability up to full workspace randomization and evaluates held-out object--location pairings that break familiar associations without increasing spatial difficulty. Across these stress tests and data scaling, we find that for representative VLA policies, including SmolVLA and $\pi_{0.5}$, execution of the manipulation primitive remains substantially more reliable than instruction-conditioned task success in harder regimes, suggesting that manipulation skill acquisition is decoupled from instruction following. We recommend augmenting manipulation benchmarks with task ladders and decomposed metrics that separately measure primitive execution and instruction-conditioned success to better diagnose instruction-grounded generalization.
\end{abstract}

%% file: icml2026/1_introduction.tex
\begin{table*}[ht!]
\centering
\caption{Success rates on LIBERO, Meta-World, and our high-randomization ManiSkill-based benchmark.
Figures for LIBERO and MetaWorld are taken from LeRobot~\cite{smolvla,lerobot}, while results on our benchmark are obtained using the same models and training code.
Performance decreases sharply as region-based layout regularities are progressively removed in our environment.}
\label{tab:main}
\begin{tabular}{llccccc}
\toprule
\textbf{Benchmark} & \textbf{Policy (\# Params)} & \multicolumn{5}{c}{\textbf{Success Rate (\%) -- Simulation}} \\
\midrule
\multirow{4}{*}{LIBERO}  & & Spatial & Object & Goal & Long & Avg. \\
 & $\pi_{0}$ ($3.3$B) & $90$ & $86$ & $95$ & $73$ & $86.0$ \\
 & $\pi_{0.5}$ ($3.6$B)  & $97$ & $99$ & $98$ & $96$ & $97.5$ \\
 & SmolVLA ($0.45$B) & $90$ & $96$ & $92$ & $71$ & $87.3$ \\
 \midrule
 \multirow{3}{*}{MetaWorld}  & & Easy & Medium & Hard & Very Hard & Avg. \\
 & $\pi_{0}$ ($3.5$B) & $71.8$ & $48.2$ & $41.7$ & $30.0$ & $47.9$ \\
 & SmolVLA ($0.45$B) & $82.5$ & $41.8$ & $45.0$ & $60.0$ & $57.3$ \\
 \midrule
\multirow{3}{*}{Ours} & & Small Jitter & Medium Jitter & Large Jitter & Full Random & Avg. \\
 & SmolVLA ($0.45$B) & $90$ & $42$ & $48$ & $2$ & $45.5$ \\
\bottomrule
\end{tabular}
\end{table*}

\section{Introduction}

Vision-language action (VLA) policies~\cite{kim2024openvla,pi05,geminirobotics15,nvidia2025gr00tn1} aim to enable general-purpose robotic behavior by integrating visual perception, language conditioning, and control. Recent work reports strong performance on robotic manipulation benchmarks~\cite{libero,rlbench,metaworld}, often achieving high success with relatively few demonstrations. These results suggest that VLA policies can acquire effective manipulation skills and generalize across a range of tasks and environments.

However, benchmark success alone does not explain \emph{why} a policy succeeds. Many widely used manipulation benchmarks rely on fixed layouts or tightly constrained object placements, introducing strong layout-dependent regularities. In such settings, policies may exploit correlations between instructions and object locations rather than learning a robust mapping from language to the intended object instance. When these regularities are weakened, performance can degrade sharply, raising concerns that standard evaluations may overestimate instruction-following ability.

We study this issue in a simple but diagnostic setting: multi-object picking. This task requires both executing a manipulation primitive and selecting the correct target object based on the instruction, in the presence of distractors. As a result, it allows us to disentangle failures of manipulation skill execution from failures of instruction-grounded target selection. We present a controlled evaluation that probes instruction grounding along two complementary axes. First, we progressively increase object placement variability, from structured placements with limited randomization to full workspace randomization, weakening layout-dependent cues. Second, we evaluate compositional generalization by holding out specific object--location pairings, where all objects and all locations are individually observed during training but their combinations are unseen at test time.

Across these stress tests, we observe a consistent pattern: policies often retain the ability to execute the manipulation primitive while increasingly failing to select the instructed target once object--location correlations are disrupted. As shown in Table~\ref{tab:main}, models that achieve strong performance in benchmark-like settings experience a sharp drop in success as region-based shortcuts are removed, a trend that is further exposed by compositional hold-out tests even under limited spatial variability.

These findings motivate a more diagnostic evaluation perspective. Rather than reporting only aggregate task success, we argue that manipulation benchmarks should incorporate task ladders that progressively weaken layout-dependent cues, together with decomposed metrics that distinguish primitive execution from instruction-following. Our main contribution is a controlled and reproducible evaluation framework for diagnosing VLA generalization using multi-object picking as an illustrative setting, showing how strong benchmark performance can mask brittle instruction-following behavior.

\subsection{Environment Randomization in VLA Evaluation}

A key factor shaping reported VLA performance is the amount of spatial structure present in benchmark environments. Many commonly used benchmarks instantiate objects within small regions around fixed canonical locations, resulting in limited variation across episodes. While this design simplifies data collection and evaluation, it also enables policies to succeed by memorizing stereotypical motions or exploiting stable location cues, rather than by robustly grounding instructions in perception.

To directly assess the role of such structure, we evaluate VLA models under progressively increased spatial randomization, up to fully random object placement. Learning instruction-grounded behavior in these settings may require substantially more data than the few hundred demonstrations per task typically used in existing benchmarks~\cite{libero,metaworld}. We therefore introduce an automated data collection pipeline based on state-based reinforcement learning agents, enabling us to generate expert datasets of up to $100{,}000$ demonstrations and test whether instruction grounding emerges through data scaling alone.

Our least randomized \emph{Small Jitter} setting closely matches the placement variability of LIBERO, and we recover comparable performance when training with a similar number of demonstrations. However, performance degrades sharply as spatial randomization increases, dropping to $2\%$ success under full workspace randomization, even when trained on $100{,}000$ demonstrations. These results indicate that increased spatial variability exposes limitations in instruction-grounded generalization that are not resolved by naive data scaling.

%% file: icml2026/2_related.tex
\section{Related Work}
\label{sec:related_work}

\paragraph{Vision-Language Action Policies.}
Vision-language action (VLA) policies~\cite{kim2024openvla,octo2024,tri2025lbm,pi0,pi05,pistar06,smolvla,geminirobotics,geminirobotics15,nvidia2025gr00tn1} integrate visual perception, language conditioning, and control to produce instruction-conditioned robotic behavior. A central motivation behind these generalist policies is that large-scale pretraining can yield reusable representations that support zero-shot or low-data adaptation to downstream manipulation tasks. In this work, we examine how such capabilities are reflected in the standard simulation benchmarks commonly used to evaluate VLA systems. We focus on two widely adopted, flow-matching-based VLA models, SmolVLA and $\pi_{0.5}$. To ensure consistency with prior evaluations, we follow standard training and evaluation practices using the community-maintained LeRobot library~\cite{lerobot}, initializing from publicly released pretrained checkpoints.

\paragraph{Manipulation Benchmarks, Generalization, and Shortcut Learning.}
Manipulation benchmarks used to evaluate VLA policies, such as LIBERO~\cite{libero}, CALVIN~\cite{calvin}, RLBench~\cite{rlbench}, and MetaWorld~\cite{metaworld}, vary in task diversity but often share strong structural regularities. Objects are typically placed within fixed or narrowly constrained regions, yielding limited spatial variability and tight coupling between object identity and location. These properties reduce the need for recombining known elements and can enable shortcut learning~\cite{shortcut_learning}, where policies exploit layout-dependent correlations or memorize stereotypical trajectories rather than robustly grounding language in perception.

As a result, high task success may conflate manipulation skill with instruction-grounded target selection. In multi-object scenes, policies can succeed by relying on stable object--location associations, masking failures such as grasping the wrong object or ignoring redundant instruction attributes. Recent work has begun to expose these issues: ~\cite{liberopro} and~\cite{liberoplus} introduce robustness stress tests that reveal sensitivity to object placement and evaluation protocol, while~\cite{vlabench} reports very low success rates on evaluations requiring generalization to unseen tasks or instructions.

Our work complements these efforts by systematically manipulating spatial variability and object--location composition during training and evaluation through a task ladder. By retraining models on increasingly randomized environments and reporting decomposed metrics that separate execution from instruction-conditioned correctness, we directly probe whether instruction grounding improves once shortcut structure is removed, and diagnose how and why performance degrades as benchmark regularities are relaxed.

%% file: icml2026/3_method.tex
\begin{figure*}[ht!]
  \centering
  \begin{subfigure}[t]{0.45\textwidth}
    \centering
    \includegraphics[width=\linewidth]{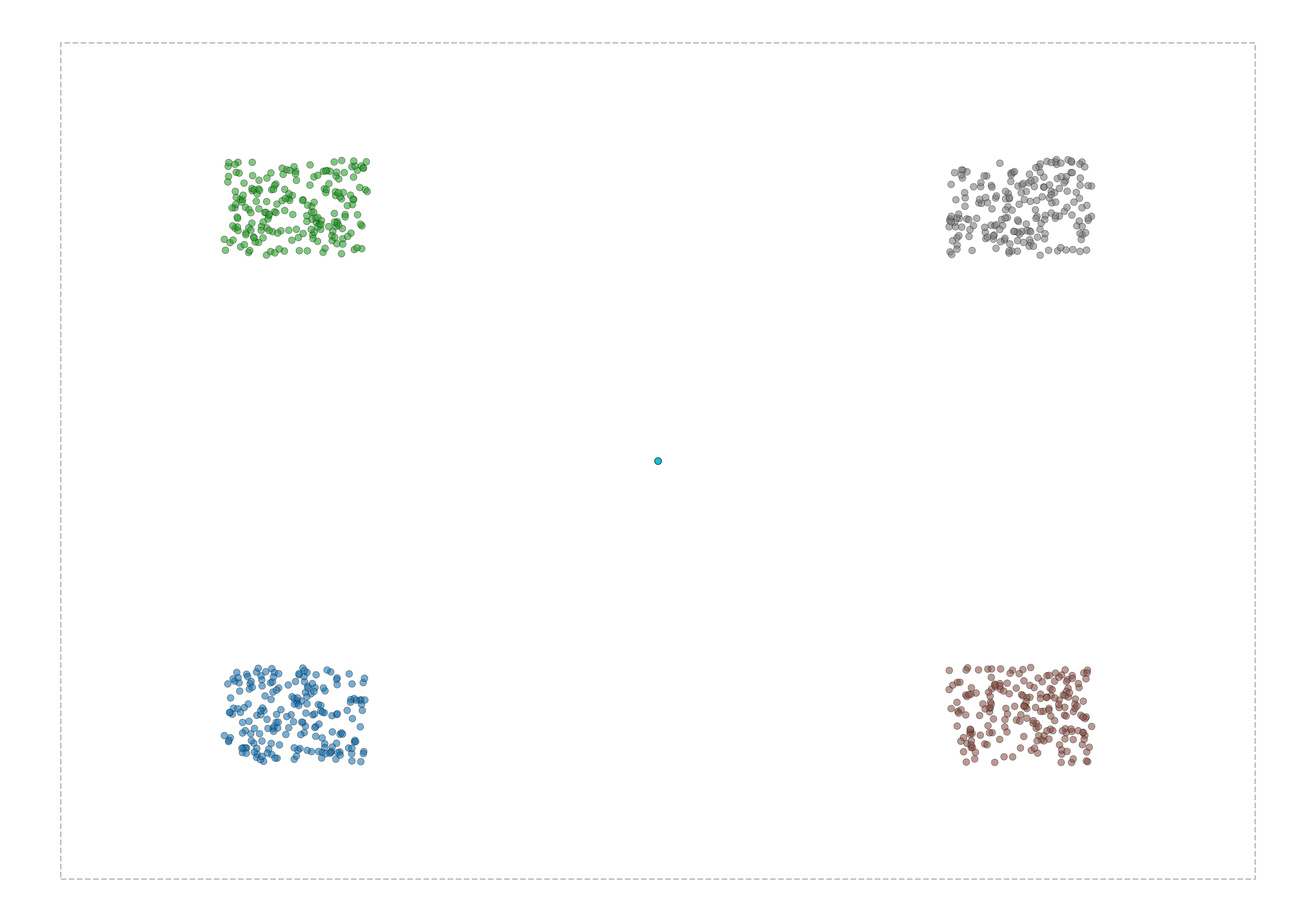}
    \caption{Small jitter}
    \label{fig:jitter_small}
  \end{subfigure}
  \hfill
  \begin{subfigure}[t]{0.45\textwidth}
    \centering
    \includegraphics[width=\linewidth]{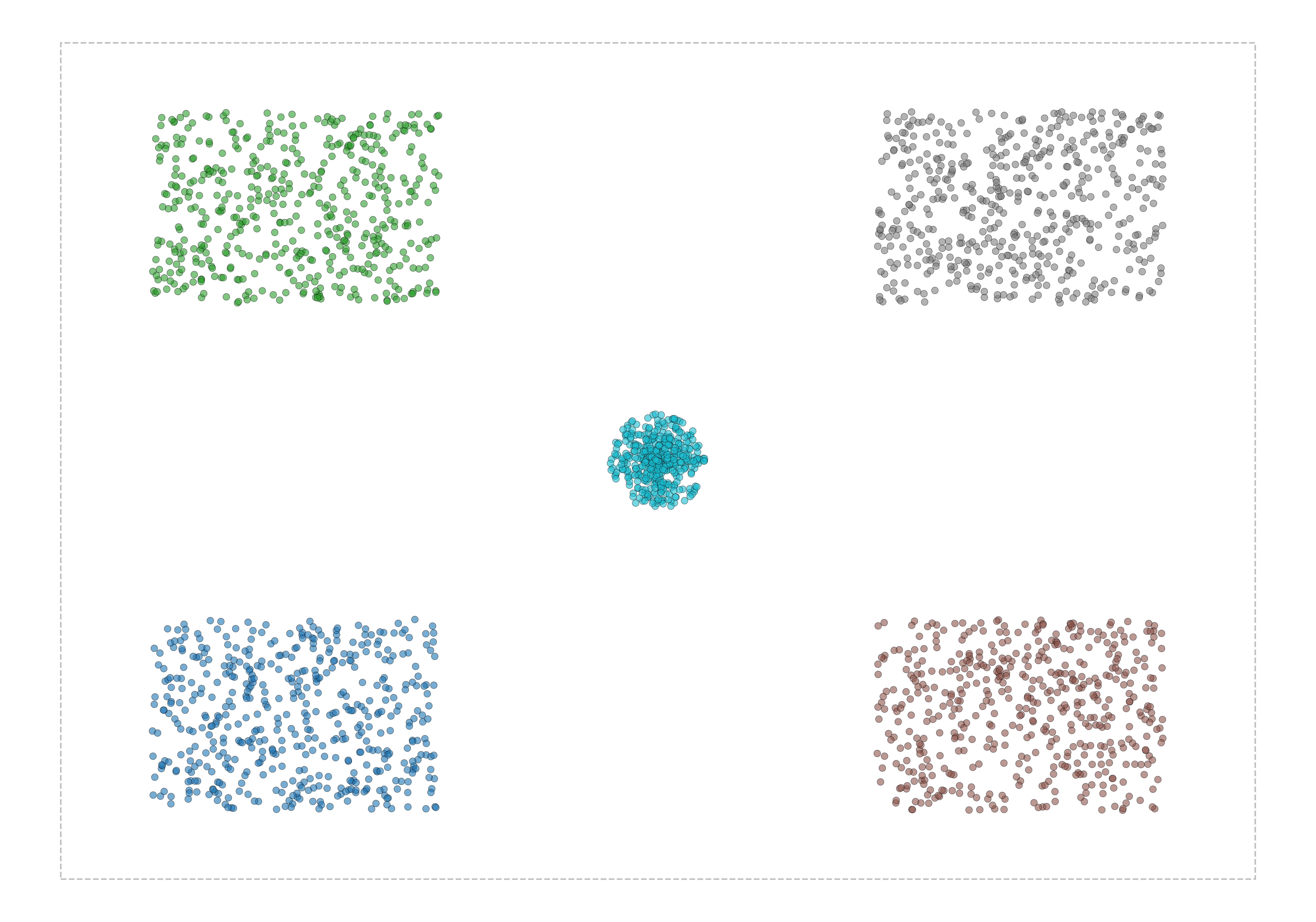}
    \caption{Medium jitter}
    \label{fig:jitter_medium}
  \end{subfigure}
  \hfill
  \begin{subfigure}[t]{0.45\textwidth}
    \centering
    \includegraphics[width=\linewidth]{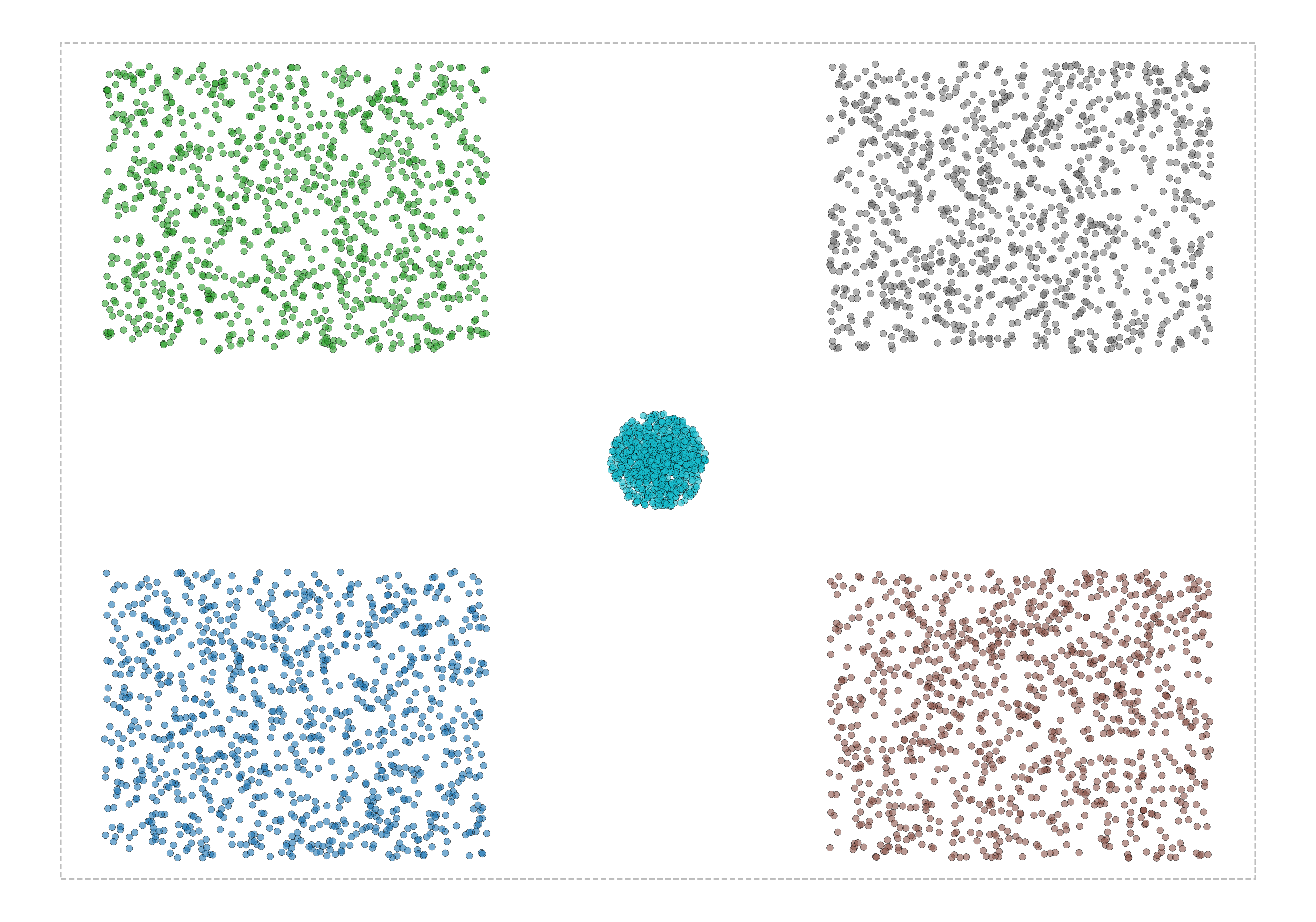}
    \caption{Large jitter}
    \label{fig:jitter_large}
  \end{subfigure}
  \hfill
  \begin{subfigure}[t]{0.45\textwidth}
    \centering
    \includegraphics[width=\linewidth]{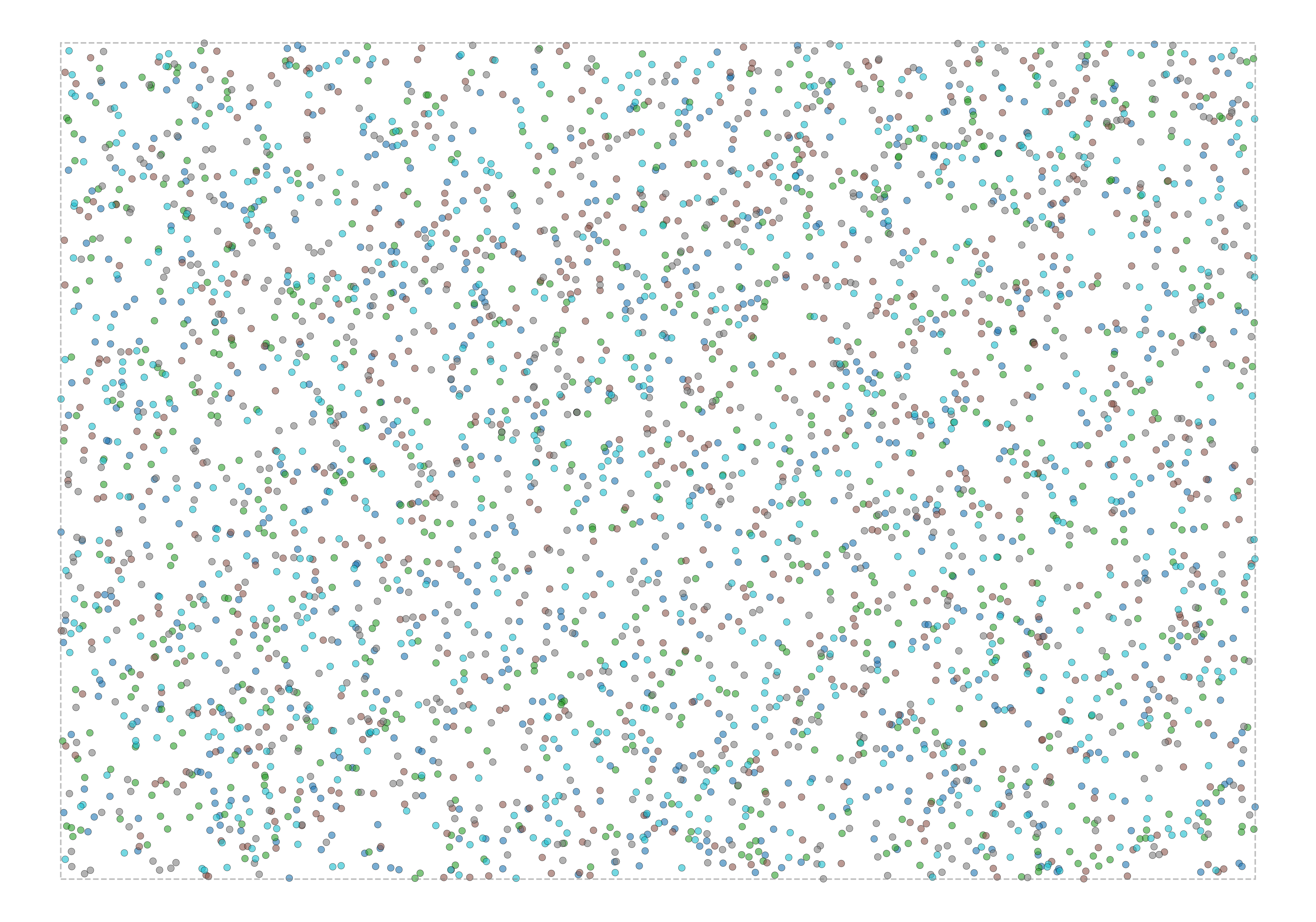}
    \caption{Full random}
    \label{fig:jitter_full_random}
  \end{subfigure}
  \caption{\textbf{Object placement distributions under increasing randomization.}
Object placements progress from fixed regions with limited jitter to full workspace randomization. Each panel shows sampled initial object positions, with colors indicating different objects. In the jittered regimes, object placements remain largely separated across regions, enabling shortcut learning based on location. In contrast, full workspace randomization produces substantial overlap between object placement distributions, eliminating reliable positional cues.}

  \label{fig:jitter_distributions_single_row}
\end{figure*}

\section{Method}
\label{sec:method}

We summarize here how we implement:
\begin{itemize}
    \item the simulated environment and its difficulty ladder,
    \item the process for generating large-scale demonstration data automatically,
    \item the training and deployment pipeline of the policies we study.
\end{itemize}

\subsection{Task Setup}
\label{subsec:task_setup}

We study a multi-object picking task involving five distinct objects placed on a tabletop. At the beginning of each episode, the agent receives a natural language instruction of the form \emph{``grasp the \textless object\textgreater.''} The task requires the agent to grasp the specified object within a fixed time horizon.

This setting is intentionally minimal yet diagnostic. Because multiple objects are present simultaneously, success requires both (i) executing a manipulation primitive (reaching and grasping) and (ii) correctly binding the language instruction to the corresponding object instance in the scene. Importantly, the manipulation primitive is identical across instructions; only the target object varies. This allows us to probe whether failures arise from poor execution or from incorrect instruction grounding.

\subsubsection{Difficulty Ladder}
\label{subsec:difficulty_ladder}

To systematically probe reliance on layout-dependent cues, we define a task ladder that progressively increases spatial variability while keeping task semantics fixed, as illustrated in Figure~\ref{fig:jitter_distributions_single_row}:
\begin{itemize}
  \item \textbf{Small jitter:} Objects are placed within small, $4\times6$cm regions around canonical workspace locations. The center object is kept fixed at the workspace center. This setting closely mirrors the limited spatial variability used in LIBERO~\cite{libero}, where pre-defined regions are $2$ to $5$cm wide. 
  \item \textbf{Medium jitter:} The size of regions is increased to $8\times12$cm. This setting falls within the range found in the MetaWorld environment~\cite{metaworld}, which defines regions of $0$ to $20$cm.
  \item \textbf{Large jitter:} Regions are further enlarged to $12\times16$cm, which aligns with the most randomized tasks from MetaWorld.
  \item \textbf{Full random:} Objects are placed uniformly across the entire $35\times50$cm workspace, subject only to collision constraints, removing region-based structure entirely.
\end{itemize}

This progression creates a smooth transition from benchmark-like configurations to highly variable environments. Across all levels, the instruction format, action space, and manipulation primitive remain unchanged, isolating the effect of spatial structure on instruction-grounded behavior.

\subsubsection{Simulation Environment}
The simulation environment and tasks are implemented in ManiSkill~\cite{maniskill}. Up to five objects from the YCB dataset~\cite{ycb} are placed on a tabletop in front of a Franka Panda robotic arm. The observations provided to the VLA policies consist of two RGB camera views—a fixed external camera offering a global view of the scene and a wrist-mounted camera providing local visual feedback during manipulation—along with proprioceptive state information, including joint positions, joint velocities, and gripper state. The action space is defined in Cartesian coordinates and consists of 6D end-effector pose deltas. A visualization of the environment is shown in Figure~\ref{fig:sim_environment}. Additional environment details and randomization settings are provided in Appendix~\ref{app:env}.

\begin{figure}[h]
  \centering
  \includegraphics[width=0.95\linewidth]{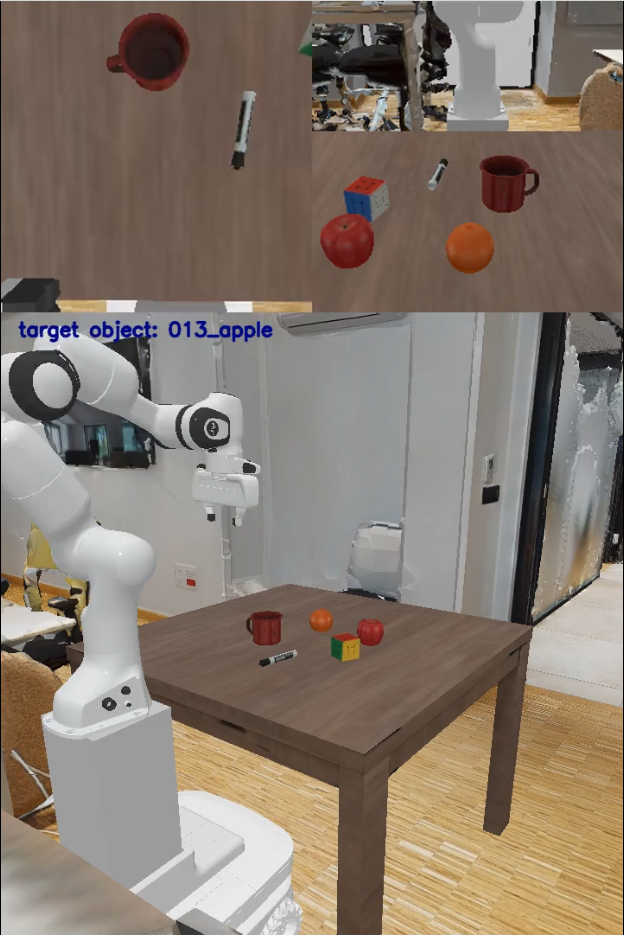}
  \caption{\textbf{Multi-object YCB environment in ManiSkill.}
The VLA observes two RGB camera streams: a wrist-mounted camera (top left) and a fixed frontal camera (top right). 
The bottom image shows the full environment for visualization only and is not provided as input to the policy.}
  \label{fig:sim_environment}
\end{figure}

\subsection{Data, training and evaluation}
\label{subsc:vlas}

\subsubsection{Demonstration Data}

We generate training data by recording state-based reinforcement learning agents trained with PPO~\cite{ppo}. These agents have access to privileged state information, including exact object poses, enabling efficient learning of high-quality grasping behavior. Agents are trained for 30 million environment steps and achieve success rates between 85\% and 98\%, depending on the number of objects present and difficulty level.

We record successful trajectories at $20$Hz for $2.5$ seconds and convert them into LeRobotDataset-v3 format~\cite{lerobot-dataset}. This automated pipeline allows us to generate datasets containing up to 100,000 demonstration trajectories, corresponding to 10 million RGB frames. Details on RL training, data filtering, and dataset construction are provided in Appendix~\ref{app:data}.

\subsubsection{VLA policies}
We evaluate two representative vision-language action policies: SmolVLA~\cite{smolvla}, a compact near-frontier model, and $\pi_{0.5}$~\cite{pi05}, a frontier-scale model. These models were selected to span a wide range of parameter counts while sharing similar training paradigms and benchmark performance profiles. Both policies are evaluated using the open-source implementations provided by LeRobot~\cite{lerobot}.

To remain representative of baseline performance, we adopt default training and inference settings from LeRobot. We perform limited grid search over key hyperparameters that are known to affect imitation learning performance, including batch size, action chunk length, and execution horizon. All hyperparameters and training details are reported in Appendix~\ref{app:implem}.

\subsection{Evaluation Metrics}
\label{subsec:metrics}

We evaluate policies using decomposed metrics designed to disentangle manipulation skill execution from instruction-grounded object selection:
\begin{itemize}
  \item \textbf{Success:} The instructed object is successfully grasped at the end of the episode.
  \item \textbf{Grasp-anything:} At least one object is grasped at any point during the episode, regardless of whether it is the instructed target.
  \item \textbf{Reach:} The end effector terminates within $5$\,cm of the instructed object at the end of the episode.
\end{itemize}

\emph{Success} reflects both correct instruction grounding and effective manipulation execution. \emph{Grasp-anything} captures the ability to execute the grasping primitive independent of target selection or final episode state, while \emph{Reach} reflects partial instruction following in cases where the policy actually approaches the correct object but fails to complete a successful grasp.

%% file: icml2026/4_experiments.tex
\section{Experiments and Results}
\label{sec:experiments}

We empirically study how increasing distribution difficulty affects what vision-language action (VLA) policies learn, with a focus on disentangling manipulation skill execution from instruction-grounded object selection.

Our experiments address three questions:
\begin{itemize}
    \item How does progressively increasing spatial variability affect manipulation skill execution versus instruction-grounded object selection?
    \item Can VLA policies generalize compositionally to unseen object--region combinations when spatial variability is controlled?
    \item Do common remedies such as increased data or reduced scene complexity mitigate failures in instruction grounding?
\end{itemize}

\paragraph{Evaluation protocol.}
All reported results are computed over 100 evaluation episodes. For each training regime, we perform a grid search over batch size, action chunk size, and execution horizon, and report the best-performing configuration.


\subsection{Spatial Variability Along the Task Ladder}
\label{subsec:exp_A}

We evaluate performance across the task ladder introduced in Section~\ref{subsec:difficulty_ladder}, which progressively increases spatial variability from benchmark-like small jitter placements to full workspace randomization.

SmolVLA is evaluated across the entire ladder, providing complete coverage of all regimes. Due to computational constraints, $\pi_{0.5}$ is evaluated only in the two most challenging and diagnostic settings: large jitter and full random placement. These regimes minimize layout-dependent shortcuts and most clearly expose failures in instruction grounding. Results are summarized in Table~\ref{tab:ladder}.

Across the ladder, SmolVLA exhibits a clear degradation in instruction-conditioned success as spatial variability increases. Success drops sharply from small jitter to medium and large jitter, though not strictly monotonically. In the jitter regimes, \emph{reach} remains consistently high, reflecting the persistence of region-based spatial structure that facilitates approximate target localization. Under full workspace randomization, however, instruction-conditioned success collapses almost entirely, indicating a failure to generalize once coarse positional cues are removed.

For $\pi_{0.5}$, performance in the large jitter regime is comparatively low, but degrades less abruptly when transitioning to full random placement. In the fully random setting, $\pi_{0.5}$ achieves higher reach and success rates than SmolVLA, suggesting that it relies on more generic behaviors that are comparatively more effective once structured region cues are fully absent.

Across both models, \emph{grasp-anything} degrades more slowly than instruction-conditioned success, while \emph{reach} remains substantially higher across all jitter levels. Taken together, these results indicate that policies retain the ability to execute coherent reach-and-grasp motions, but increasingly fail to bind those motions to the instructed object as spatial uncertainty grows. Qualitative inspection confirms this pattern: in most failures, agents execute smooth and confident grasping motions toward seemingly arbitrary locations, occasionally contacting and grasping an object by chance. Representative trajectories are shown in Appendix~\ref{app:videos}.

\begin{table}[t]
  \centering
  \caption{\textbf{Performance along the spatial task ladder.}
  The ladder progressively increases spatial variability from benchmark-like small jitter to full workspace randomization.}
  \label{tab:ladder}
  \footnotesize
  \renewcommand{\arraystretch}{1.25}
  \begin{tabular}{llcccc}
    \toprule
    Model & Regime 
    & Success 
    & Grasp-any 
    & Reach \\
    \midrule
    \multirow{4}{*}{SmolVLA}
      & Small jitter  & $90$ & $98$ & $100$ \\
      & Medium jitter & $37$ & $48$ & $93$ \\
      & Large jitter  & $41$ & $50$ & $100$ \\
      & Full random   & $2$  & $12$ & $4$ \\
    \midrule
    \multirow{2}{*}{$\pi_{0.5}$}
      & Large jitter  & $10$ & $11$ & $60$ \\
      & Full random   & $6$  & $15$ & $16$ \\
    \bottomrule
  \end{tabular}
\end{table}


\subsection{Compositional Generalization to Held-Out Object-Region Pairings}
\label{subsec:exp_B}

Finally, we evaluate compositional generalization by holding out specific object-region pairings during training and evaluating them only at test time. For each object, three of the five regions are used during training, while the remaining two are reserved for evaluation. All objects and all regions are therefore observed during training, but their combinations are not.

To isolate compositional effects from spatial variability, we conduct this evaluation under the small jitter regime using SmolVLA. Results are reported in Table~\ref{tab:pairings}.

Instruction-conditioned success drops from $44\%$ on in-distribution pairings to $0\%$ on held-out pairings, indicating a complete failure to recombine known objects and regions. Decomposed metrics clarify the failure mode: \emph{reach} is perfect on in-distribution pairings but collapses on held-out ones, showing that the policy no longer reliably approaches the instructed target once learned object-location associations are broken. \emph{Grasp-anything} decreases more moderately, indicating that grasp execution remains partially intact but is no longer guided by the instruction.

These results show that even in benchmark-like settings with limited spatial variability, current VLA policies struggle with compositional generalization once positional shortcuts are removed.

\begin{table}[t]
  \centering
  \caption{\textbf{Compositional generalization to held-out object--region pairings.}
  SmolVLA is trained under the small-jitter regime with each object appearing in three regions.
  Performance is evaluated on seen object--region pairings (ID) and on the two held-out regions per object (OOD).}
  \label{tab:pairings}
  \footnotesize
  \renewcommand{\arraystretch}{1.25}
  \begin{tabular*}{\columnwidth}{@{\extracolsep{\fill}}lccc@{}}
    \toprule
    Split & Success & Grasp-any & Reach \\
    \midrule
    Pairings (ID)  & $44$ & $57$ & $100$ \\
    Pairings (OOD) & $0$  & $31$ & $2$ \\
    \bottomrule
  \end{tabular*}
\end{table}


\subsection{Ablations: Data Scaling and Scene Complexity}
\label{subsec:ablations}

Next, we examine whether failures under increased spatial variability can be mitigated by scaling training data or simplifying the scene.

\subsubsection{Data scaling}

We evaluate whether increasing the number of demonstrations mitigates failures observed under higher spatial variability by scaling dataset size in the medium jitter, large jitter, and full random regimes. Results are reported in Table~\ref{tab:data_scaling}.

In the medium and large jitter regimes, increasing the dataset size leads to modest improvements for both SmolVLA and $\pi_{0.5}$ across metrics. Success and grasp-anything increase, while reach remains largely saturated. These gains indicate that additional data can improve execution robustness in intermediate regimes, although object selection relies primarily on region cues rather than precise instruction following.

In the full random regime, where spatial structure is entirely removed, scaling the dataset from $10{,}000$ to $100{,}000$ trajectories yields no meaningful improvement for SmolVLA across any metric. In contrast, $\pi_{0.5}$ shows small but consistent gains in reach, grasp-anything, and success. While absolute success remains low, this makes $\pi_{0.5}$ the stronger model in the most randomized setting.

This difference suggests that, under extreme spatial uncertainty, model scale and capacity may confer some advantage in maintaining weak instruction-conditioned behavior. However, even at this scale, increased data alone is insufficient to recover robust instruction grounding once layout-dependent cues are removed.

Overall, these results show that data scaling can modestly strengthen execution-oriented behavior, and may benefit larger models in the hardest regimes, but does not resolve the dominant failure mode under high spatial uncertainty.

\begin{table}[h]
  \centering
  \caption{\textbf{Response to data scaling under increasing spatial variability.}
  Values report metric changes as dataset size increases.
  Med.\ jitter: $0.5\mathrm{k}\!\rightarrow\!1\mathrm{k}$,
  Lrg.\ jitter: $1\mathrm{k}\!\rightarrow\!5\mathrm{k}$,
  Full random: $10\mathrm{k}\!\rightarrow\!100\mathrm{k}$.}
  \label{tab:data_scaling}
  \footnotesize
  \renewcommand{\arraystretch}{1.15}
  \begin{tabular*}{\columnwidth}{@{\extracolsep{\fill}}llccc@{}}
    \toprule
    Model & Regime & Success & Grasp-any & Reach \\
    \midrule
    \multirow{3}{*}{SmolVLA}
      & Med. jitter  & $37\!\rightarrow\!42$  & $48\!\rightarrow\!54$  & $93\!\rightarrow\!100$ \\
      & Lrg. jitter  & $41\!\rightarrow\!48$ & $50\!\rightarrow\!55$ & $100\!\rightarrow\!100$ \\
      & Full random  & $2\!\rightarrow\!1$   & $12\!\rightarrow\!10$ & $4\!\rightarrow\!4$ \\
    \midrule
    \multirow{2}{*}{$\pi_{0.5}$}
      & Lrg. jitter  & $10\!\rightarrow\!18$ & $11\!\rightarrow\!22$ & $60\!\rightarrow\!67$ \\
      & Full random  & $4\!\rightarrow\!6$   & $13\!\rightarrow\!15$ & $11\!\rightarrow\!16$ \\
    \bottomrule
  \end{tabular*}
\end{table}

\subsubsection{Effect of object count}

To test whether simplifying the scene mitigates instruction-following failures, we reduce the number of objects in the full random environment. Results for SmolVLA are reported in Table~\ref{tab:object_count}.

Reducing the number of objects from five to two does not improve instruction-conditioned success and in fact lowers grasp-anything rates. We hypothesize that with fewer objects, the policy is less likely to incidentally contact and grasp an object when executing a task-agnostic reach, reducing apparent execution success.

In contrast, when only a single object is present, removing the need for instruction-conditioned selection, success increases substantially despite fully random placement. This confirms that the model can learn a generalizable manipulation primitive in isolation, while highlighting that failures in multi-object settings stem from target selection rather than execution. Qualitatively, single-object behavior is less stereotyped and less smooth, consistent with the increase in reach relative to success.

\begin{table}[t]
  \centering
  \caption{\textbf{Object-count ablation under full random placement.}
  Results are obtained with a SmolVLA model trained on $10{,}000$ trajectories.}
  \label{tab:object_count}
  \footnotesize
  \renewcommand{\arraystretch}{1.25}
  \begin{tabular*}{\columnwidth}{@{\extracolsep{\fill}}ccccc@{}}
    \toprule
    \# Objects & Success  & Grasp-any  & Reach \\
    \midrule
    $1$ & $15$ & $24$ & $38$ \\
    $2$ & $2$ & $4$ & $4$ \\
    $3$ & $1$ & $3$ & $6$ \\
    $4$ & $2$ & $7$ & $4$ \\
    $5$ & $2$ & $12$ & $4$ \\
    \bottomrule
  \end{tabular*}
\end{table}

%% file: icml2026/5_discussion.tex
\section{Discussion}
\label{sec:discussion}

Across all experiments, we observe a clear separation between manipulation skill execution and instruction-grounded object selection as positional structure is weakened. Vision-language action (VLA) policies consistently retain the ability to execute coherent grasping motions, yet increasingly fail to select the instructed object once layout-dependent cues are removed or disrupted. This separation emerges along two complementary axes: spatial variability and compositional generalization. Together, these experiments show that strong manipulation performance can coexist with brittle instruction grounding, even in settings that closely resemble standard benchmark conditions.

Along the spatial task ladder, instruction-conditioned success depends strongly on positional structure. In benchmark-like regimes with limited randomization, policies achieve high success rates, consistent with prior benchmark results. As spatial variability increases within the jitter regimes, success degrades while reach remains high, indicating continued reliance on coarse region-level cues to approach the correct area. In these settings, manipulation skill remains largely intact and can be modestly improved with additional data. In contrast, under full workspace randomization—where region-based shortcuts are eliminated—both reach and success collapse, despite non-trivial grasp execution. This marks a transition from execution-limited behavior under structured layouts to failures of instruction-conditioned target selection once positional cues are unavailable.

The compositional generalization experiment reveals an even sharper failure mode. When object--region pairings are held out during training, instruction-conditioned success drops from $44\%$ on in-distribution pairings to $0\%$ on held-out ones, despite all objects and all regions being individually observed. Decomposed metrics show that reach collapses alongside success, while grasp-anything remains partially intact. Notably, this failure occurs under the small jitter regime, where spatial variability and control difficulty are minimal. This demonstrates that even in benchmark-like settings, policies struggle to recombine known objects and locations once learned object--location correlations are broken.

Ablation results reinforce this interpretation. Increasing the amount of demonstration data yields only modest gains in execution-oriented metrics under intermediate variability and no meaningful improvement in instruction-conditioned success under full randomization, even with an order-of-magnitude increase in data. Similarly, reducing scene complexity does not restore instruction grounding in multi-object settings, with success improving only when instruction-conditioned selection is no longer required. These findings indicate that the dominant limitation is not data scarcity or control instability, but a breakdown in instruction-conditioned target selection once shortcut structure is removed.

Overall,  our findings show that strong benchmark performance does not necessarily imply robust instruction-conditioned generalization. Aggregate task success alone can obscure failures in instruction grounding. Evaluations with decomposed metrics that systematically weaken layout-dependent shortcuts and test compositional generalization provide a more diagnostic view of what VLA policies learn.

%% file: icml2026/6_conclusion.tex
\section{Limitations, Conclusion, and Future Work}
\label{sec:limitations_conclusion}

\subsection{Limitations}
This work is deliberately diagnostic rather than prescriptive. We focus on a single manipulation primitive in a controlled simulated environment and do not propose new architectures, training objectives, or grounding mechanisms. While this simplicity enables clean isolation of instruction-grounded failure modes, it limits the scope of our conclusions. We examine a restricted set of distributional challenge: spatial object placement variability and compositional object--region generalization. We do not address other sources of difficulty such as linguistic paraphrasing, appearance variation, long-horizon reasoning, or real-world embodiment. Additionally, although we evaluate both a compact and a frontier-scale policy, coverage across regimes is not uniform, and results should be interpreted in light of this experimental scope.

\subsection{Conclusion}
We presented a controlled evaluation framework for diagnosing restricted generalization in vision-language action policies using a multi-object picking task. By progressively weakening layout-dependent structure and evaluating held-out object--location pairings, we showed that reliable execution of a manipulation primitive can coexist with brittle instruction-grounded object selection. Across models, regimes, and ablations, failures in harder settings consistently reflect breakdowns in language-to-instance binding rather than deficiencies in control or data scale. These results demonstrate that high success rates in structured benchmarks can mask instruction-grounding failures that only become visible once positional shortcuts are removed.

\subsection{Future Work}
Future work should explore architectural and training approaches that more explicitly support instruction-conditioned target selection under spatial variability. Promising directions include language-conditioned, object-centric representations; mechanisms that preserve instruction influence throughout execution; and learning signals that penalize incorrect target selection even when manipulation succeeds. Extending the proposed evaluation framework to additional manipulation primitives, longer-horizon tasks, and more diverse environments would further test the generality of these findings. More broadly, incorporating task ladders, compositional generalization tests, and decomposed metrics into benchmark design may enable more faithful measurement of instruction-grounded generalization in VLAs.

%% file: icml2026/7_appendix.tex
\section{VLA implementation details}
\label{app:implem}

\subsubsection*{Models and checkpoints}
We use LeRobot checkpoints: \texttt{lerobot/pi05\_base} for pi0.5 and \texttt{lerobot/smolvla\_base} for smolVLA. For pi0.5 we set \texttt{dtype=bfloat16}, \texttt{gradient\_checkpointing=true}, \texttt{freeze\_vision\_encoder=true}, and \texttt{train\_expert\_only=true}; smolVLA uses LeRobot defaults. Both policies take two RGB views (\texttt{observation.images.camera1}, \texttt{observation.images.camera2}, shape $[3{,}256{,}256]$) and a state vector (\texttt{observation.state}, shape $[15]$).

\subsubsection*{Training}
Default training uses 30\,000 steps, batch size 128, and chunk size 16 for full random regime. In the jitter settings as well as the compositional object-region pairings we use batch size 64.

\subsubsection*{Grid search}
We sweep batch size (e.g.\ 64, 128), chunk size (e.g.\ 8, 16, 32), and also the policy execution horizon (e.g.\ 1, 4, 8, 16). Evaluation uses the same \texttt{chunk\_size} as training and, in the best configs from our sweep for execution horizon. We typically found this to be 4 or 8.

Our best grid search results in full random were:
smolvla: chunk size=32, batch size=128, execution horizon = 8
pi05: chunk size=32, batch size=128, execution horizon = 8

\section{Environment Specifications}
\label{app:env}

\subsubsection*{Simulation and Robot}
All experiments are conducted in ManiSkill with GPU-accelerated simulation. We implement a custom multi-object tabletop task, \texttt{GraspObject-v1}, using a Franka Panda robotic arm. The robot is controlled in Cartesian space using a proportional-derivative end-effector delta pose controller (\texttt{pd\_ee\_delta\_pose}). Actions are executed at 20\,Hz, while the underlying physics simulation runs at 100\,Hz.

\subsubsection*{Objects and Layout}
Graspable objects are drawn from the YCB object set. We use a fixed subset of five objects: apple, orange, Rubik’s cube, mug, and large marker. Under jittered placement regimes, each object appears within a predefined spatial region of the workspace, with increasing randomization across regimes, as illustrated in Figure~\ref{fig:jitter_distributions_single_row}. In the full random regime, objects are placed uniformly at random across the tabletop using a rejection sampler to avoid collisions, as shown in Figure~\ref{fig:jitter_full_random}.

\subsubsection*{Observations}
The policy observes two RGB camera streams: a wrist-mounted camera and a fixed frontal camera, both rendered at a resolution of $256\times256$. The wrist-mounted and frontal cameras have fields of view of $58^\circ$ and $45^\circ$, respectively. Proprioceptive observations consist of a 15-dimensional state vector including joint positions (7), joint velocities (7), and gripper width (1).

\section{RL training and demonstration recording}
\label{app:data}
\subsubsection*{RL expert training}
We employ the following key hyperparameters to train our RL expert which is used to generate the demos. The full list can be found in Table~\ref{tab:rl_hyperparams}.

\begin{table}[h!]
  \centering
  \label{tab:rl_hyperparams}
  \caption{\textbf{RL expert training hyperparameters.}
  State-based PPO agent trained with privileged information.}
  \label{tab:rl_hyperparams}
  \footnotesize
  \renewcommand{\arraystretch}{1.15}
  \begin{tabular*}{\columnwidth}{@{\extracolsep{\fill}}ll@{}}
    \toprule
    \textbf{Category} & \textbf{Value} \\
    \midrule
    Total timesteps        & $30{,}000{,}000$ \\
    Parallel environments  & $1024$ \\
    Max episode steps      & $50$ \\
    \midrule
    Optimizer              & AdamW \\
    Learning rate          & $1\times10^{-4}$ \\
    Weight decay           & $1\times10^{-5}$ \\
    \midrule
    Rollout steps          & $50$ \\
    Minibatches            & $2$ \\
    Update epochs          & $4$ \\
    Discount factor $\gamma$ & $0.8$ \\
    GAE parameter $\lambda$  & $0.9$ \\
    PPO clip coefficient   & $0.2$ \\
    Entropy coefficient    & $0.1$ \\
    \bottomrule
  \end{tabular*}
\end{table}

\subsubsection*{Recording pipeline}
We implement a client server architecture across LeRobot and Maniskill to record our datasets.
The server runs the trained PPO in the sim and serves steps over WebSockets. The client connects, receives full episodes, and writes only successful ones to a LeRobot dataset.

\subsubsection*{Dataset format}
Episodes are saved in LeRobotDataset format with \texttt{observation.images.hand\_camera}, \texttt{observation.images.fixed\_camera} (256$\times$256 RGB), \texttt{observation.state} (float32, dim.\ 15), \texttt{action} (float32, dim.\ 7), \texttt{task} string, \texttt{robot\_type=panda}, and standard FPS/episode/frame fields. We collect up to $\sim$100\,k successful trajectories ($\sim$10M camera images) by repeated runs with resume.


\section{Examples of produced trajectories}
\label{app:videos}

\begin{figure}[htbp]
    \centering
    \begin{subfigure}{0.32\textwidth}
        \centering
        \includegraphics[width=\textwidth, keepaspectratio]{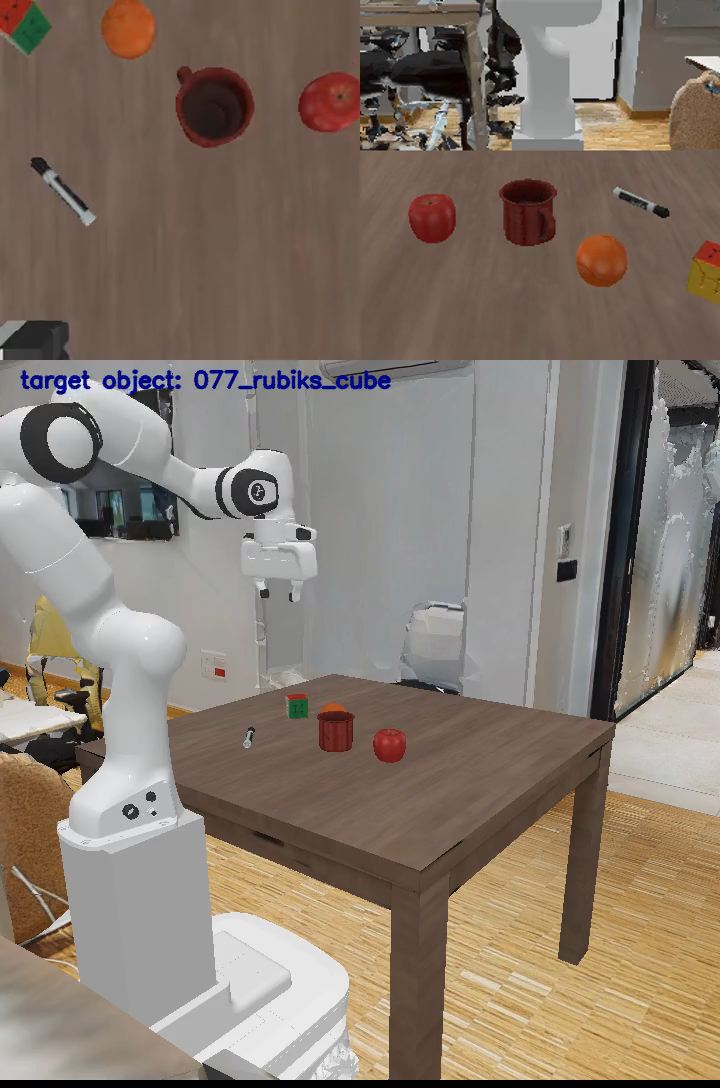}
        \caption{$t = 0$}
        \label{fig:frame1}
    \end{subfigure}
    \hfill
    \begin{subfigure}{0.32\textwidth}
        \centering
        \includegraphics[width=\textwidth, keepaspectratio]{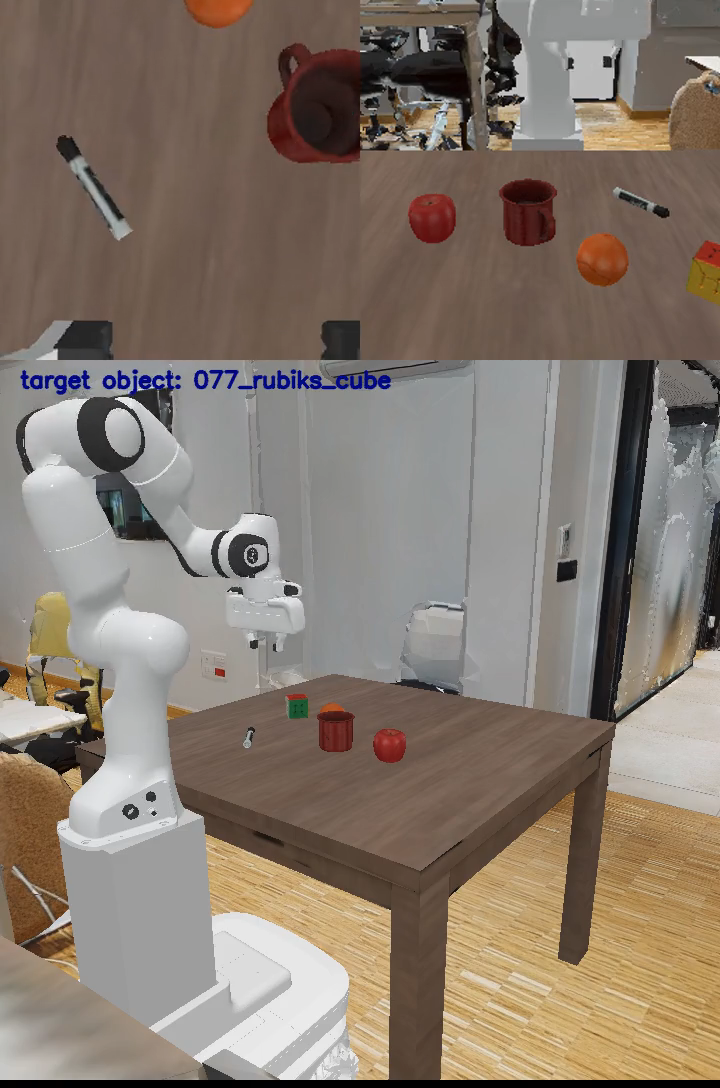}
        \caption{$t = 300\mathrm{ms}$}
        \label{fig:frame2}
    \end{subfigure}
    \hfill
    \begin{subfigure}{0.32\textwidth}
        \centering
        \includegraphics[width=\textwidth, keepaspectratio]{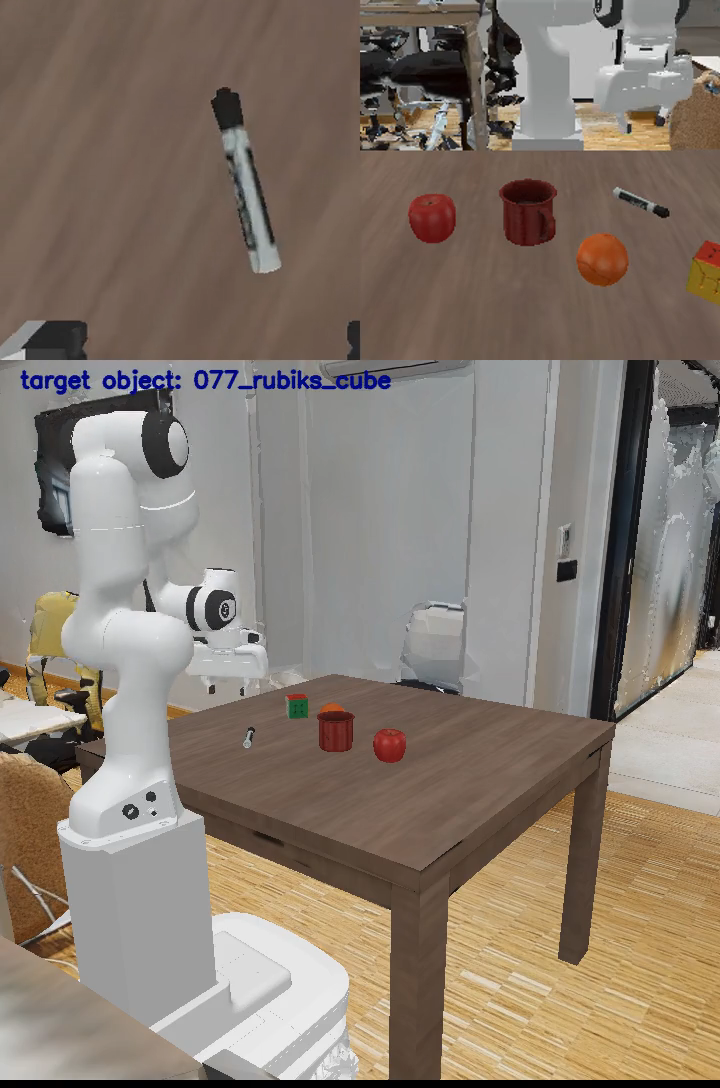}
        \caption{$t = 600\mathrm{ms}$}
        \label{fig:frame3}
    \end{subfigure}

    \vspace{0.5em} 

    \begin{subfigure}{0.32\textwidth}
        \centering
        \includegraphics[width=\textwidth, keepaspectratio]{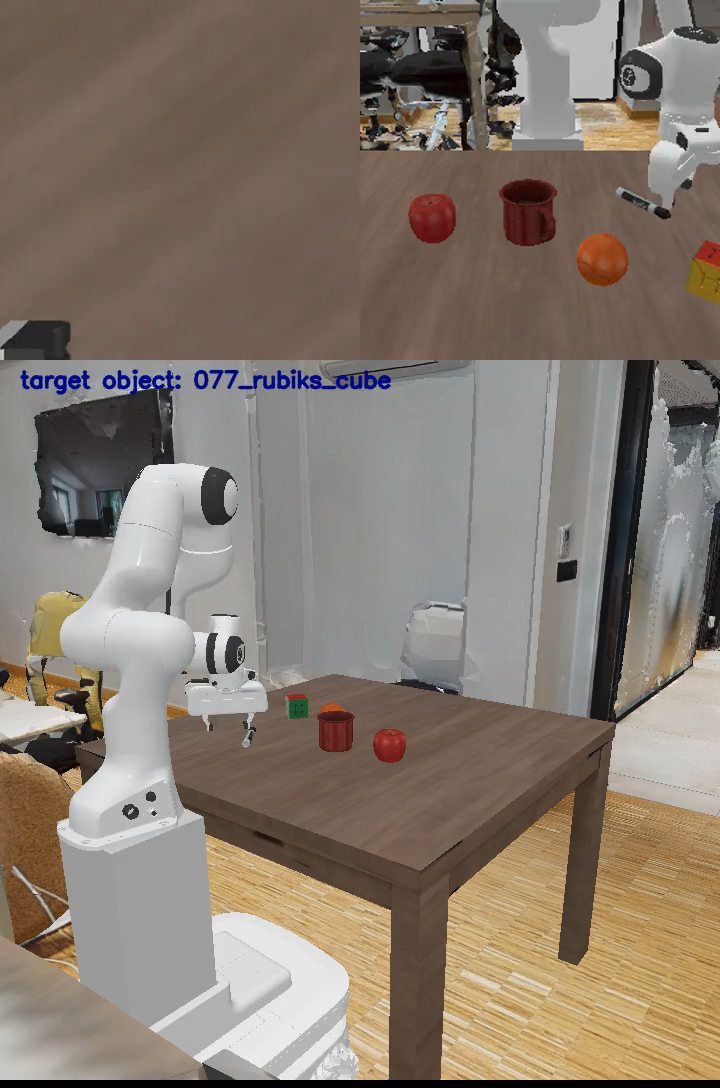}
        \caption{$t = 900\mathrm{ms}$}
        \label{fig:frame4}
    \end{subfigure}
    \hfill
    \begin{subfigure}{0.32\textwidth}
        \centering
        \includegraphics[width=\textwidth, keepaspectratio]{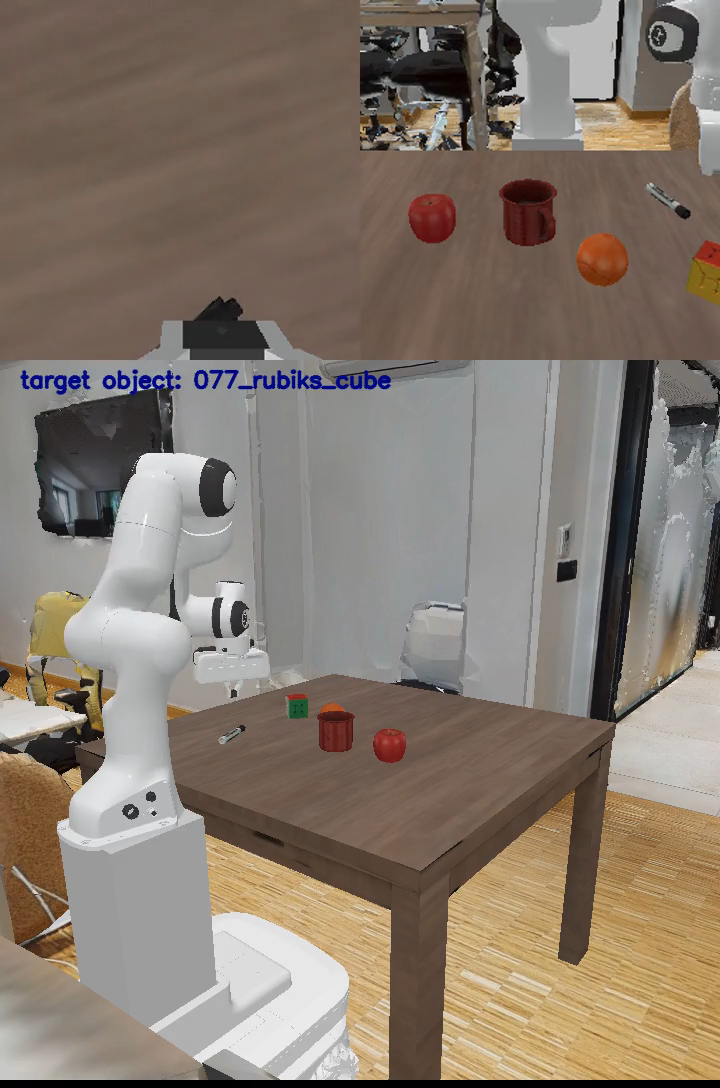}
        \caption{$t = 1200\mathrm{ms}$}
        \label{fig:frame5}
    \end{subfigure}
    \hfill
    \begin{subfigure}{0.32\textwidth}
        \centering
        \includegraphics[width=\textwidth, keepaspectratio]{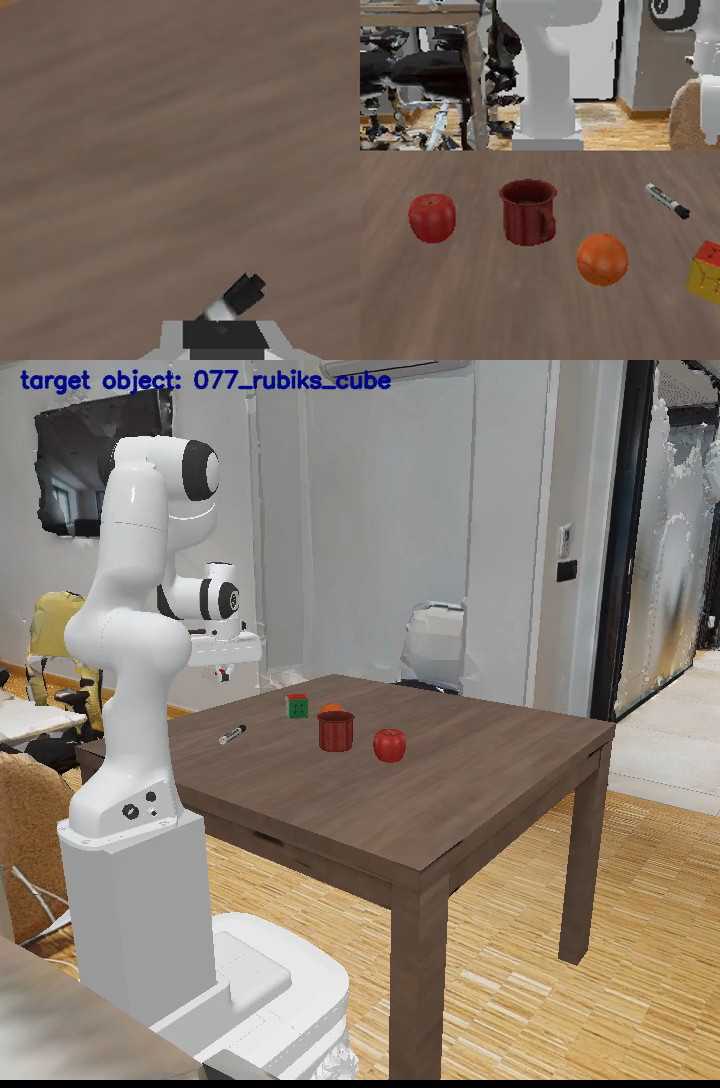}
        \caption{$t = 1500\mathrm{ms}$}
        \label{fig:frame6}
    \end{subfigure}
    \caption{Example of a trajectory produced by a $\pi_{0.5}$ model trained on $100k$ expert demonstrations in the Full Random environment.}
    \label{fig:video-frames-1}
\end{figure}

\begin{figure}[htbp]
    \centering
    \begin{subfigure}{0.32\textwidth}
        \centering
        \includegraphics[width=\textwidth, keepaspectratio]{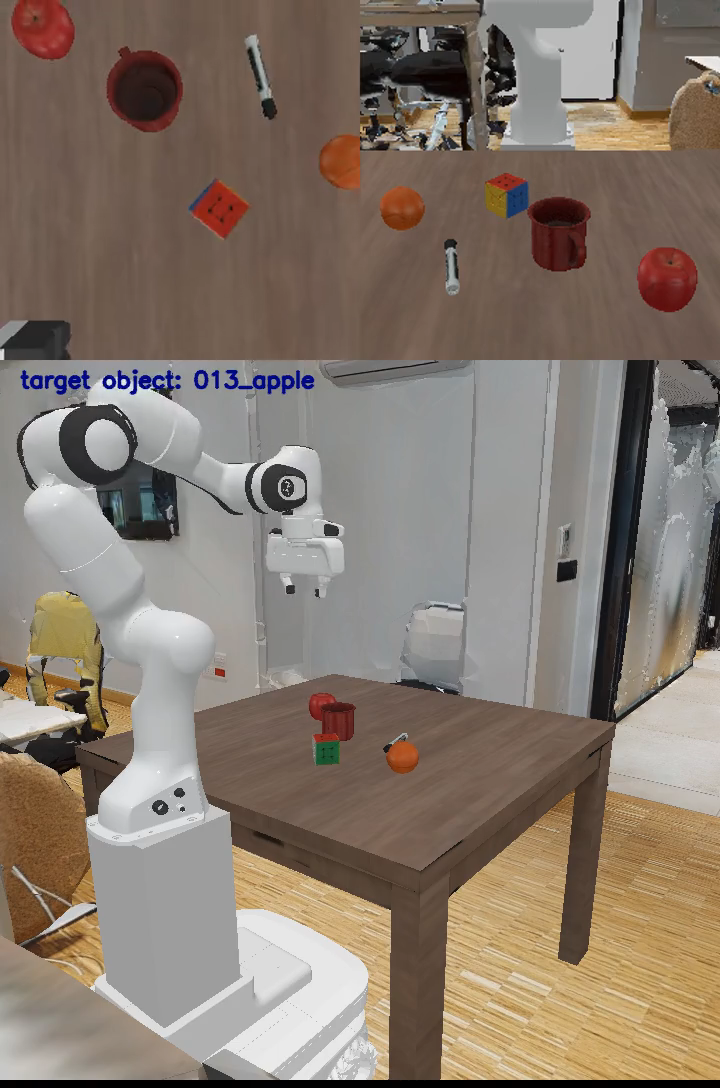}
        \caption{$t = 0$}
        \label{fig:frame1}
    \end{subfigure}
    \hfill
    \begin{subfigure}{0.32\textwidth}
        \centering
        \includegraphics[width=\textwidth, keepaspectratio]{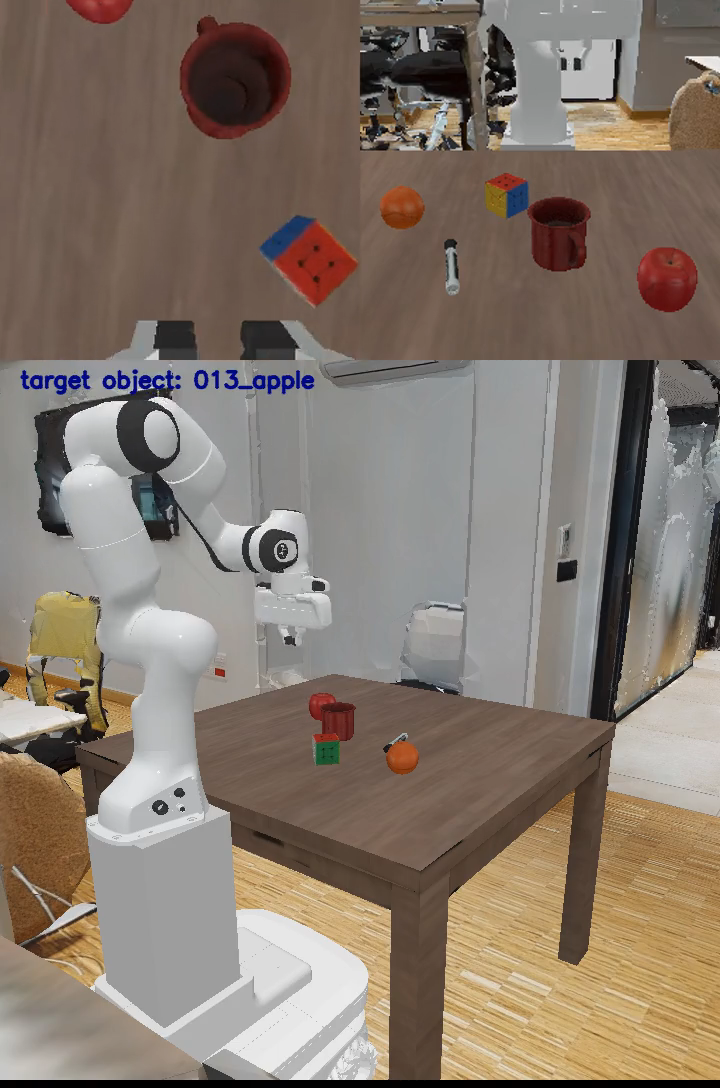}
        \caption{$t = 300\mathrm{ms}$}
        \label{fig:frame2}
    \end{subfigure}
    \hfill
    \begin{subfigure}{0.32\textwidth}
        \centering
        \includegraphics[width=\textwidth, keepaspectratio]{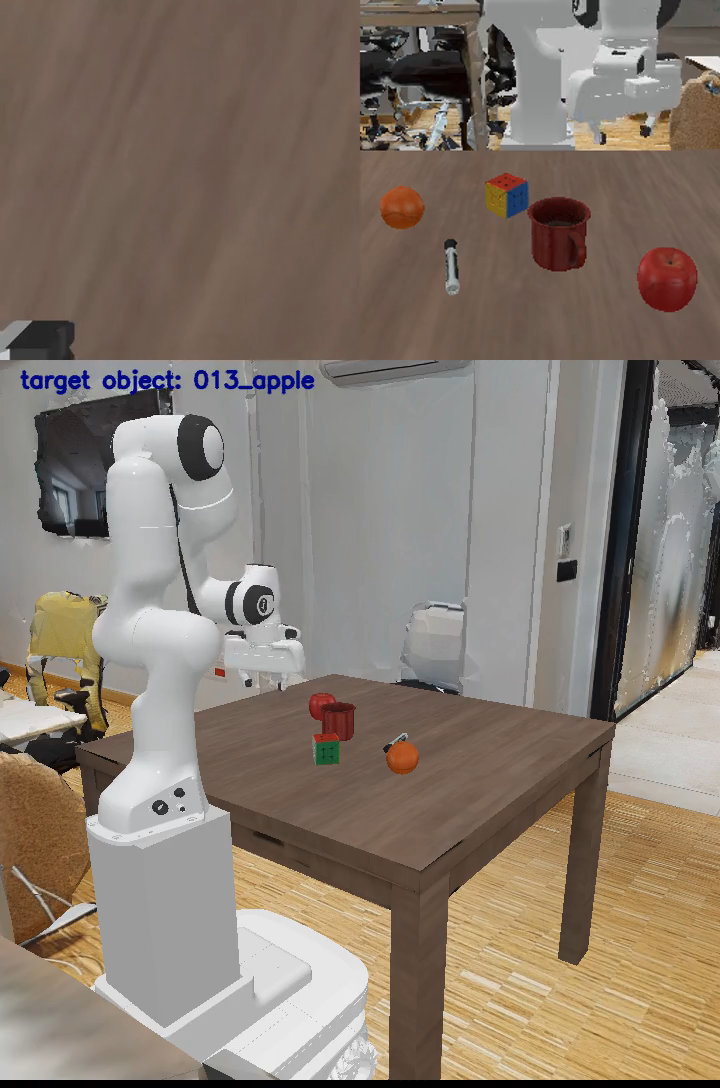}
        \caption{$t = 600\mathrm{ms}$}
        \label{fig:frame3}
    \end{subfigure}

    \vspace{0.5em} 

    \begin{subfigure}{0.32\textwidth}
        \centering
        \includegraphics[width=\textwidth, keepaspectratio]{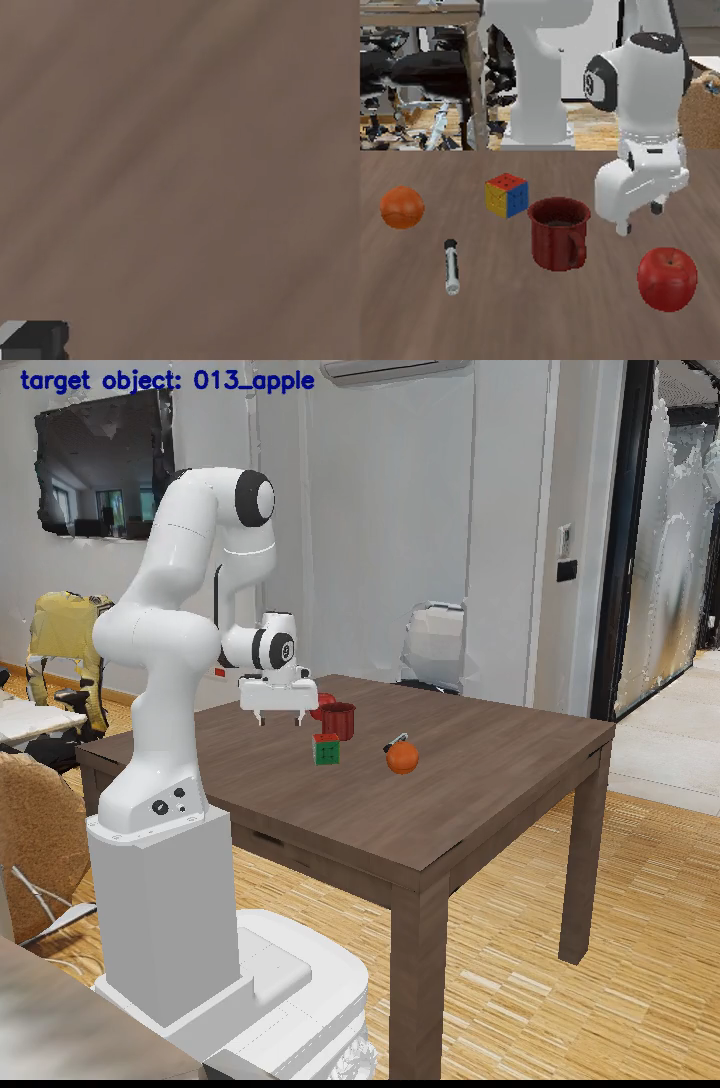}
        \caption{$t = 900\mathrm{ms}$}
        \label{fig:frame4}
    \end{subfigure}
    \hfill
    \begin{subfigure}{0.32\textwidth}
        \centering
        \includegraphics[width=\textwidth, keepaspectratio]{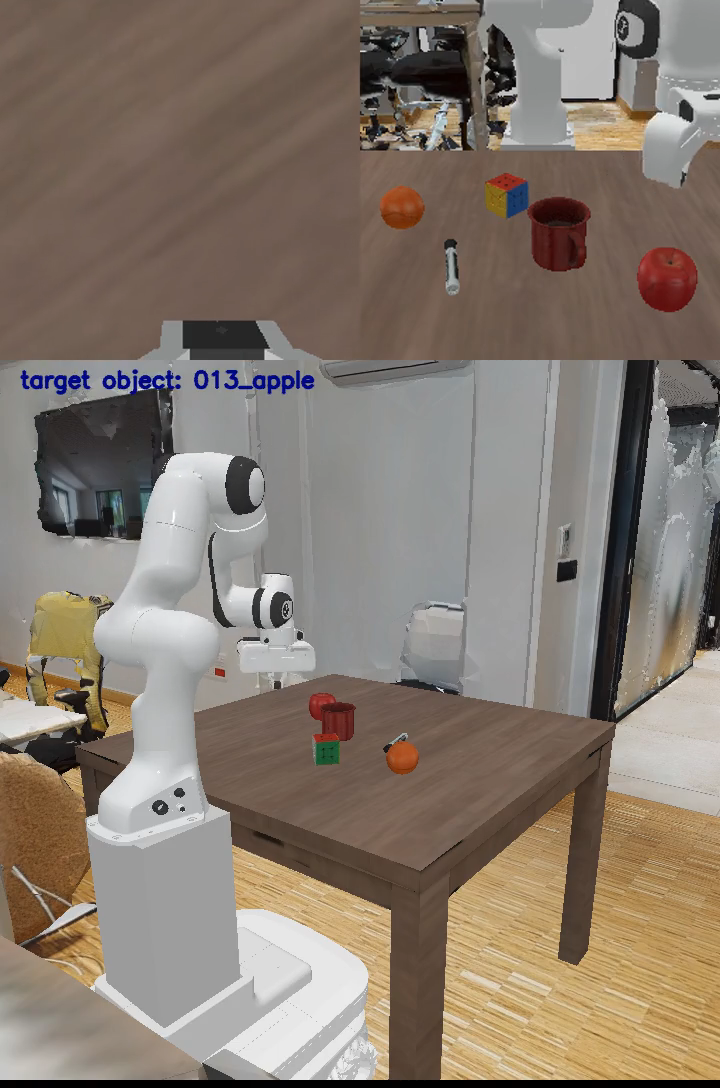}
        \caption{$t = 1200\mathrm{ms}$}
        \label{fig:frame5}
    \end{subfigure}
    \hfill
    \begin{subfigure}{0.32\textwidth}
        \centering
        \includegraphics[width=\textwidth, keepaspectratio]{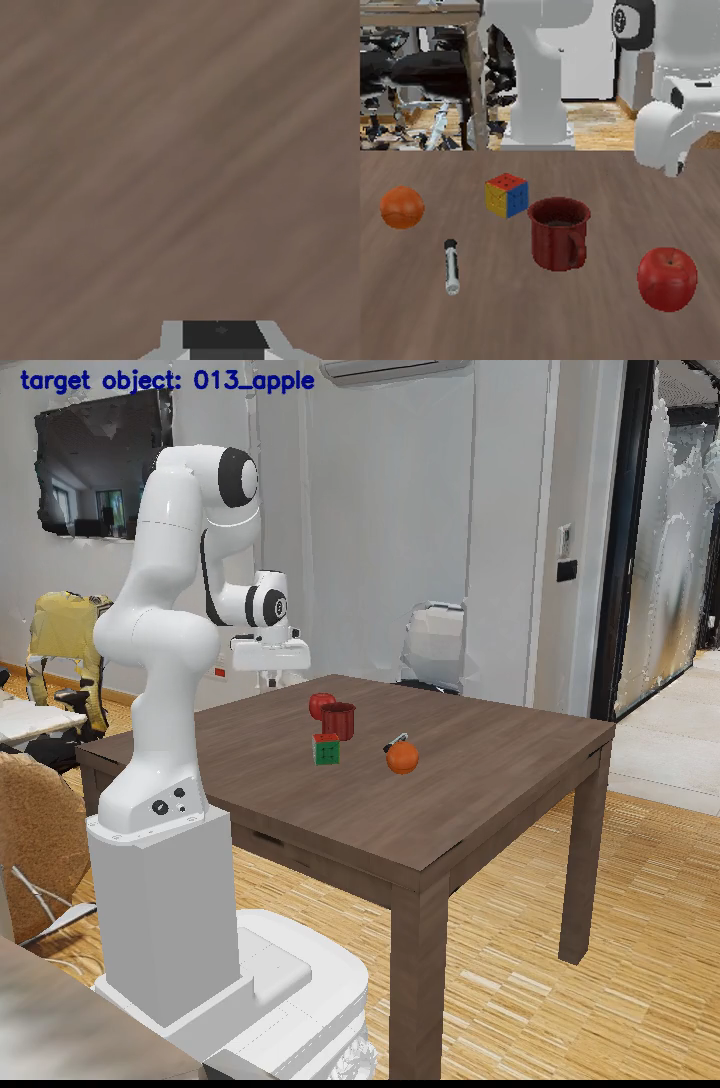}
        \caption{$t = 1500\mathrm{ms}$}
        \label{fig:frame6}
    \end{subfigure}
    \caption{Example of a trajectory produced by a $\pi_{0.5}$ model trained on $100k$ expert demonstrations in the Full Random environment.}
    \label{fig:video-frames-2}
\end{figure}